%% file: main.tex
\pdfoutput=1
\documentclass[11pt]{article}

\usepackage[final]{acl}
\usepackage{times}
\usepackage{latexsym}

\usepackage[T1]{fontenc}
\usepackage[utf8]{inputenc}

\usepackage{microtype}

\usepackage{inconsolata}

\usepackage{graphicx}

\usepackage{floatrow}
\usepackage{subcaption}
\usepackage{listings}
\usepackage{longtable}
\usepackage{bm} 
\usepackage{multirow}
\usepackage{multicol}
\usepackage{algorithm, algorithmic}
\usepackage{amsmath}
\usepackage{amssymb}
\usepackage{booktabs}
\usepackage{listings}
\usepackage{marvosym} 

\usepackage[nameinlink,capitalise]{cleveref}
\crefname{section}{§}{§§}
\Crefname{section}{§}{§§}
\crefname{lemma}{lemma}{lemmas}
\Crefname{lemma}{Lemma}{Lemmas}
\crefname{thm}{theorem}{theorems}
\Crefname{thm}{Theorem}{Theorems}

\input{custom}

\title{Generative Psycho-Lexical Approach for Constructing\\Value Systems in Large Language Models}

\author{
  Haoran Ye\thanks{Equal contribution.}\textsuperscript{,1}, Tianze Zhang\footnotemark[1]\textsuperscript{,1,2},
  Yuhang Xie\footnotemark[1]\textsuperscript{,1}, \\
  \textbf{Liyuan Zhang\textsuperscript{1},
  Yuanyi Ren\textsuperscript{1},  
  Xin Zhang\textsuperscript{3,4}, Guojie Song\thanks{Corresponding author.}\textsuperscript{,1,5}}
  \\[0.5em]
\textsuperscript{1}State Key Laboratory of General Artificial Intelligence,\\School of Intelligence Science and Technology, Peking University\\
\textsuperscript{2}Yuanpei College, Peking University\\
\textsuperscript{3}School of Psychological and Cognitive Sciences, Peking University\\
\textsuperscript{4}Key Laboratory of Machine Perception (Ministry of Education), Peking University\\
\textsuperscript{5}PKU-Wuhan Institute for Artificial Intelligence\\[0.5em]
\small \texttt{\{hrye, ericzhang, yuhangxie, zly2003\}@stu.pku.edu.cn} \\
\small \texttt{
\{yyren, zhang.x, gjsong\}@pku.edu.cn} 
}

\begin{document}

\maketitle

\input{sections/00_abstract}
\input{sections/01_introduction}
\input{sections/02_related_work}
\input{sections/03_approach}
\input{sections/04_experiements}

\input{sections/05_conclusion}

\section*{Acknowledgements}
This work is supported by the State Key Laboratory of General Artificial Intelligence; National Natural Science Foundation of China (Grant No. 62276006); Wuhan East Lake High-Tech Development Zone National Comprehensive Experimental Base for Governance of Intelligent Society; and the Fundamental Research Funds for the Central Universities.

\bibliography{main}

\clearpage
\appendix

\input{appendix/experimental_details}
\input{appendix/extended_results}
\input{appendix/against_bhn}

\end{document}

%% file: custom.tex
\usepackage[colorinlistoftodos,prependcaption,textsize=tiny]{todonotes}
\usepackage{xargs}

\newcommand{\our}{\texttt{GPLA}}
\newcommandx{\HY}[2][1=]{\todo[linecolor=orange,backgroundcolor=orange!25,bordercolor=orange,#1]{HY: #2}}

\definecolor{promptbackground}{rgb}{0.98, 0.98, 0.98} % Lighter background for prompts
\definecolor{textcolor}{rgb}{0, 0, 0} % Darker text color for better readability
\definecolor{backcolour}{rgb}{0.95, 0.95, 0.92}
\definecolor{codegreen}{rgb}{0,0.6,0}
\definecolor{codegray}{rgb}{0.5,0.5,0.5}
\definecolor{codepurple}{rgb}{0.58,0,0.82}
\definecolor{codemagenta}{rgb}{0.78,0,0.52}
\definecolor{codeblue}{rgb}{0.1,0.1,0.9}

\lstdefinestyle{promptstyle}{
    backgroundcolor=\color{promptbackground},
    basicstyle=\tiny\color{textcolor}, % Italic and footnotesize for text snippets
    breakatwhitespace=true,         % Only break lines at whitespaces
    breaklines=true,                % Enable line breaking
    captionpos=b,                   % Position caption at the bottom
    keepspaces=true,                % Preserve spaces for alignment
    showspaces=false,               % Don't show spaces explicitly
    showstringspaces=false,         % Don't show spaces in strings
    showtabs=false,                 % Don't show tabs
    frame=single,                   % Single frame around the text
    rulecolor=\color{textcolor},    % Use text color for frame
    framesep=3pt,                   % Padding between text and frame
    frameround=tttt,                % All corners rounded
    framexleftmargin=5pt,           % Extra margin on the left for symmetry
    xleftmargin=5pt,                % Left margin for alignment
    xrightmargin=5pt,               % Right margin for alignment
    tabsize=2,                      % Standard tab size
    linewidth=\textwidth            % Line width fit to page
}

\lstdefinestyle{promptcodestyle}{
    backgroundcolor=\color{promptbackground},
    commentstyle=\color{codegreen},
    keywordstyle=\color{codemagenta}, % Changed from magenta to custom magenta
    numberstyle=\tiny\color{codegray},
    stringstyle=\color{codepurple},
    basicstyle=\tiny\ttfamily, % Changed from tiny to footnotesize for better readability
    breakatwhitespace=false,
    breaklines=true,
    captionpos=b,
    keepspaces=true,
    numbersep=8pt, % Increased separation between numbers and text for clarity
    showspaces=false,
    showstringspaces=false,
    showtabs=false,
    tabsize=2, % Increased tab size for better alignment
    frame=single, % Adds a frame around the code
    frameround=tttt,                % All corners rounded
    rulecolor=\color{black}, % Color of the frame
}

\usepackage[nameinlink,capitalise]{cleveref}
\crefname{section}{§}{§§}
\Crefname{section}{§}{§§}
\crefname{lemma}{lemma}{lemmas}
\Crefname{lemma}{Lemma}{Lemmas}
\crefname{thm}{theorem}{theorems}
\Crefname{thm}{Theorem}{Theorems}

\usepackage{makecell}

%% file: sections/00_abstract.tex
\begin{abstract}
Values are core drivers of individual and collective perception, cognition, and behavior.
Value systems, such as Schwartz's Theory of Basic Human Values, delineate the hierarchy and interplay among these values, enabling cross-disciplinary investigations into decision-making and societal dynamics.
Recently, the rise of Large Language Models (LLMs) has raised concerns regarding their elusive intrinsic values. 
Despite growing efforts in evaluating, understanding, and aligning LLM values, a psychologically grounded LLM value system remains underexplored.
This study addresses the gap by introducing the Generative Psycho-Lexical Approach (\our{}), a scalable, adaptable, and theoretically informed method for constructing value systems. Leveraging \our{}, we propose a psychologically grounded five-factor value system tailored for LLMs. 
For systematic validation, we present three benchmarking tasks that integrate psychological principles with cutting-edge AI priorities.
Our results reveal that the proposed value system meets standard psychological criteria, better captures LLM values, improves LLM safety prediction, and enhances LLM alignment, when compared to the canonical Schwartz's values.
\end{abstract}

%% file: sections/01_introduction.tex
\section{Introduction}

Personal values are broad, fundamental motivations behind individual and collective actions. They serve as guiding principles that shape perceptions, cognition, and behaviors across situations and over time. As Large Language Models (LLMs) continue to permeate and transform human society, researchers have increasingly sought to evaluate, understand, and align LLM values \citep{ye2025gpv, ren2024valuebench, rao2024normad, meadows2024localvaluebench, kovavc2024stick, jiang2024raising, rozen2024llms, yao2024clave, sorensen2024value}. Rather than relying solely on specific risk metrics, implicit preference modeling, or atomic ethical principles, framing these studies through the lens of intrinsic values \cite{schwartz2012overview} offers a more comprehensive, adaptable, and transparent approach \cite{yao2023value_fulcra, biedma2024beyond, yao2023alignment_goals}. 

However, value systems used in existing research—primarily Schwartz's values—are established for humans.
A psychologically informed value system tailored for LLMs remains largely unexplored, obscuring a structured and holistic perspective on their values.
A preliminary attempt at constructing LLM value systems \cite{biedma2024beyond} has several limitations that undermine its psychological grounding; we present related discussion and experimental evidence in \cref{sec:human_to_llm_values} and \cref{app:against_bhn}.

To address this gap, we introduce a generative psycho-lexical approach (\our{}), a novel methodology that leverages LLMs to construct a value system grounded in psychological principles. This approach is based on the psycho-lexical hypothesis, which suggests that all salient values are captured in language. \our{} adopts an agentic framework and involves five sequential steps: 1) extracting perceptions from the corpus, 2) identifying values behind perceptions, 3) filtering values, 4) non-reactive value measurements of test subjects, and 5) structuring the value system. In contrast to traditional methods that manually compile value lexicons and design questionnaires to collect self-report data \cite{aavik2002structure, crectu2012psycho, saucier2000isms}, \our{} automates the entire process and features two key advantages.

First, \our{} supports \textbf{agent-specific construction}. Rather than relying on value lexicons derived from human-compiled dictionaries, it utilizes LLMs to identify values within free-form agent-generated text. This principled approach facilitates the agent- and context-specific collection of value lexicons and ensures better coverage (\cref{app:comprehensiveness}).
Second, \our{} enables \textbf{nonreactive value measurement}. Traditional methods that rely on self-report are demonstrated to be unreliable and biased for humans \cite{ponizovskiy2020development} and AI \cite{dominguez2023questioning, rottger2024political, ye2025gpv}. Instead, \our{} measures values within unstructured responses to arbitrary prompts, which mitigates response biases and enhances scalability and validity \cite{ye2025gpv}.

In applying \our{} to 33 LLMs, each steered with 21 different value-anchoring prompts \cite{rozen2024llms}, we develop a five-factor value system encompassing \textit{Social Responsibility}, \textit{Risk-taking}, \textit{Rule-following}, \textit{Self-competence}, and \textit{Rationality}. We demonstrate the system's reliability and explore its theoretical and practical implications. To benchmark LLM value systems, we propose three tasks based on psychometric principles \cite{devon2007psychometric} and AI priorities \cite{anwar2024foundational}: \textit{Confirmatory Factor Analysis} for assessing structural validity, \textit{LLM Safety Prediction} for examining predictive validity, and \textit{LLM Value Alignment} for evaluating representational power. These tasks confirm the validity and utility of our system in evaluating, understanding, and aligning the intrinsic values of LLMs, when compared to canonical Schwartz's values \cite{schwartz2012overview}.

In summary, we contribute: 1) the Generative Psycho-Lexical Approach (\our{}), a novel LLM- and data-driven methodology for constructing psychologically grounded value systems; 2) a five-factor value system constructed using \our{} for LLMs, with demonstrated reliability and an exploration of its theoretical and practical implications; and 3) three benchmarking tasks for LLM value systems, through which we establish the superior validity and utility of the proposed system.

%% file: sections/02_related_work.tex
\section{Background and Related Work}

\subsection{Values, Value Measurement, and Theory-Driven Value Systems}

\paragraph{Values.}
Values are broad, fundamental beliefs and guiding principles, serving as motivational forces when evaluating actions, people, and events \cite{schwartz2012overview}. Unlike other psychological constructs such as personalities, values are stable, motivational, and transituational \cite{sagiv2022personal}, rendering them uniquely powerful across disciplines for investigating decision-making and social dynamics \cite{sen1986social, schwartz2006theory}. 

\paragraph{Value Measurement.}
Values are measurable constructs \cite{schwartz2001extending}. Value measurement seeks a quantitative assessment of the importance an agent assigns to specific values. Traditionally, it relies on tools like self-report questionnaires \cite{maio2010mental}, behavioral observation \cite{lonnqvist2013personal}, and experimental techniques \cite{murphy2014social}, which are hindered by response biases, resource demands, and inabilities to capture authentic behaviors \cite{ponizovskiy2020development, boyd2015values}. Recently proposed Generative Psychometrics for Values (GPV) \cite{ye2025gpv} leverage LLMs to dynamically parse unstructured texts into perceptions and measure values therein. GPV is shown to be more scalable, valid, and flexible than traditional tools in measuring LLM values, and is utilized for non-reactive value measurement in this work.

\paragraph{Theory-Driven Value Systems.}
Isolated values are structured into a value system for a holistic understanding of their hierarchy and interplay \cite{schwartz2012overview}. Theory-driven value systems like Schwartz's theory of basic human values \cite{schwartz1992universals} and functional theory of values \cite{gouveia2014functional} are rooted in theoretical hypotheses and preconceptions of psychologists. These systems are established prior to data collection and analysis, and are later validated through empirical studies. However, theory-driven approaches are constrained by the subjectivity of the theorists, limited coverage, and inadaptability to evolving values or specific contexts \cite{ponizovskiy2020development, raad2017reply}.

\subsection{Psycho-Lexical Approach}
In contrast to the theory-driven approaches, the psycho-lexical approach organizes psychological constructs such as values, personality traits, and social attitudes \cite{aavik2002structure, crectu2012psycho, Klages1929-KLATSO-5, saucier2000isms}, under a data-driven paradigm. It operates on the fundamental principle that language naturally evolves to capture salient and socially relevant individual differences \cite{de2016values}.
The pioneering attempts date back to the early 20th century \cite{allport1936trait, galton1950character}, and the paradigm has then been extended to different psychological constructs
in decades of development \cite{cattell1957personality, john1988lexical, aavik2002structure,ponizovskiy2020development, mai2023exploring, garrashi2024personality}.

Most of these works, taking values as an example, involve the following steps: 1) compilation of value-descriptive terms, mostly from dictionaries and thesauruses; 2) refinement and reduction to eliminate redundant, ambiguous, and uncommon terms; 3) value measurement, through collecting self-reports and peer ratings, to identify underlying correlations between the value descriptors; and 4) uncovering the hidden value factors or dimensions through statistical methods like principal component analysis or factor analysis.

In this work, we aim to harness the extensive knowledge and semantic understanding of LLMs to address the limitations of traditional psycho-lexical approaches. These limitations include: 1) extensive manual labor required to compile and refine the lexicons, 2) insufficient coverage of values of different linguistic forms, 3) lack of criteria for prioritizing values lexicons when filtering, 4) the responses bias and inscalability of self-report value measurement, and 5) the inadaptability to changing values or specific contexts.

\subsection{From Human Values to LLM Values}
\label{sec:human_to_llm_values}
The rise of LLMs also brings the critical need to evaluate, understand, and align their values \citep{ye2025large}. Extensive works focus on evaluating the values of LLMs using traditional self-report tools \cite{li2022gpt,li2024evaluatingpsychologicalsafetylarge, huang2024humanity, safdari2023-personality-traits-in-llm, safdari2023personality}, static customized inventories \cite{ren2024valuebench,meadows2024localvaluebench, jiang2024evaluating}, dynamically generated inventories \cite{jiang2024raising}, or directly from the model's free-form outputs \cite{ye2025gpv, yao2024clave,yao2023value_fulcra, yao2025leaderboard}.
Other works attempt to understand the LLM values in aspects like the consistency of shown values \cite{rozen2024llms, moore2024large,rottger2024political}, the ability to reason about human values \cite{ganesan2023systematic,sorensen2024value, jiang2024can, strachan2024testing}, and the ability to express or role-play human values \cite{jiang-etal-2024-personallm, zhang2023measuring, kang2024causal, zhang2025extrapolating, li2024quantifying, huang2024social}.
Another line of research aligns LLM values with human values, setting the alignment goal as specific risk metrics \cite{lin2024towards, gunjal2024detecting}, human demonstrations \cite{dubois2024alpacafarm, kopf2024openassistant, alpaca}, implicit preference modeling \cite{ouyang2022training, rafailov2024dpo, zhong2024panacea}, or intrinsic values \cite{bai2022hh, yao2023value_fulcra, bai2022constitutional}. Among different alignment goals, intrinsic values, as transituational decision-guiding principles, demonstrate unique advantages \cite{yao2023alignment_goals}. However, prior works are mostly based on Schwartz's value system, which is established and validated for humans and may not necessarily capture the unique LLM psychology.

One preliminary attempt \cite{biedma2024beyond} at constructing LLM value systems faces several limitations that undermine its psychological grounding: 1) the value lexicon collection is susceptible to response bias and lacks comprehensive coverage; 2) the derivation of value correlations lacks theoretical or empirical validity; and 3) the resulting value system has yet to be evaluated in terms of its reliability, validity, or utility. \cref{app:against_bhn} presents further discussion and experimental evidence.

%% file: sections/03_approach.tex
\section{Generative Psycho-Lexical Approach for Constructing LLM Value System}
\label{sec:approach}

\begin{figure*}[t]
  \centering
  \begin{subfigure}{\textwidth}
    \centering
    \includegraphics[width=\linewidth]{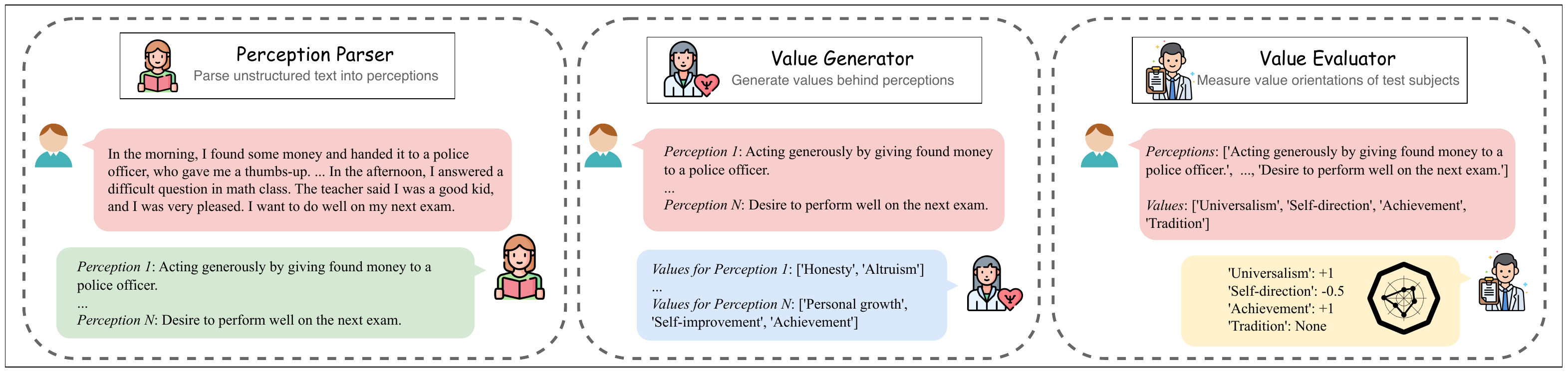}
    \caption{LLM agents in \our{}: (1) Perception Parser $M_{P}$, (2) Value Generator $M_{G}$, and (3) Value Evaluator $M_{E}$.}
    \label{fig:sub1}
  \end{subfigure}
  
  \begin{subfigure}{\textwidth}
    \centering
    \includegraphics[width=\linewidth]{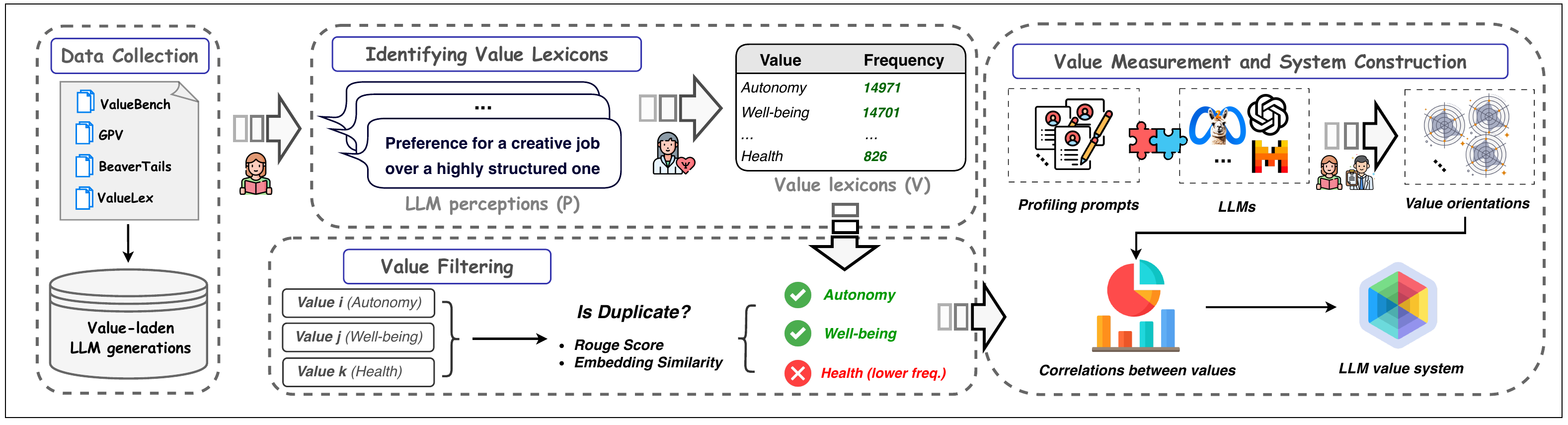}
    \caption{The pipeline of \our{} for constructing LLM value system: 1) extracting perceptions from the corpus, 2) identifying values behind perceptions, 3) filtering values, 4) measuring value orientations of LLM test subjects, and 5) computing the structure of the value system.}
    \label{fig:sub2}
  \end{subfigure}
  
  \caption{Generative Psycho-lexical Approach (\our{}) for Constructing LLM Value System.}
  \label{fig:test}
\end{figure*}

Generative Psycho-Lexical Approach (\our{}), as our foundation for constructing the LLM value system, is illustrated schematically in \cref{fig:test} and algorithmically in \cref{alg:approach}. \our{} adopts an agentic framework, comprising three LLM agents and four sequential steps, to systematically collect, filter, measure, and structure the LLM value system.

\begin{algorithm*}
  \caption{Generative Psycho-lexical Approach for Constructing LLM Value System}
  \label{alg:approach}
  \begin{algorithmic}[1]
  \STATE \textbf{Input:} Corpus $\mathcal{C}$ of LLM-generated text, LLMs as value measurement subjects $M = \{M_i\}_{i=1}^{n}$, and three LLM agents $M_{P}$, $M_{G}$, $M_{E}$.
  
  \STATE \textbf{Output:} Value system $\mathcal{VS} = (V, V_H, \mathbf{\Lambda})$ of LLMs.
  
  \STATE $P = M_{P}(\mathcal{C})$ \COMMENT{Extract perceptions $P$ from corpus $\mathcal{C}$}
  \STATE $\{V, Freq\} = M_{G}(P)$ \COMMENT{Generate values $V$ behind perceptions $P$ and collect their frequencies}
  \STATE $V = \text{LexiconFiltering}(V, Freq)$ \COMMENT{Remove duplicate and less prioritized values}
  \STATE $\mathbf{X}_{M} = \text{GPV}(V, M; M_{P}, M_{E})$  \COMMENT{Measure the value orientations of LLM subjects}
  \STATE $V_H, \mathbf{\Lambda} = \text{PCA}(\mathbf{X}_{M})$ \COMMENT{Compute the hidden factors and structure of the system}
  \end{algorithmic}
\end{algorithm*}

\paragraph{Data Collection.} In line with the established psycho-lexical approach \cite{de2016values}, we begin by comprehensively identifying atomic values that LLMs may endorse. We draw upon a diverse, value-laden corpus of LLM outputs \(\mathcal{C}\) from ValueBench \cite{ren2024valuebench}, GPV \cite{ye2025gpv}, BeaverTails \cite{ji2024beavertails}, and ValueLex \cite{biedma2024beyond}. These corpora provide a broad spectrum of value-laden LLM generations. Detailed statistics and further explanations on these datasets are provided in \cref{app:data_statistics}.

\paragraph{Identifying Value Lexicons.} Then, we extract perceptions \( P \) from the collected corpus using the Perception Parser \( M_{P} \). Perceptions are atomic value-rich expressions that reflect values behind free-form LLM outputs and are akin to stimuli in traditional psychometric tools \cite{ye2025gpv}. These perceptions are then mapped to underlying values \( V \) through the Value Generator \( M_{G} \), with their frequency of occurrence \( Freq \) recorded. We instantiate \( M_{P} \) following GPV \cite{ye2025gpv} and \( M_{G} \) using Kaleido model \cite{sorensen2024value}. In \cref{app:comprehensiveness}, we confirm the comprehensive coverage of the generated values.

\paragraph{Value Filtering.} To ensure the value lexicons are concise and representative, we filter out duplicates using Rouge scores and embedding similarity following the validated setup in \cite{sorensen2024value}. When two values exhibit significant semantic overlap, we retain the one more frequently observed, as higher frequency signifies greater priority \cite{ponizovskiy2020development}. This process results in value lexicons \( V \) that minimize redundancy while preserving semantic coverage.

\paragraph{Value Measurement and System Construction.}
Structuring the LLM value system requires
identifying correlations between values based on LLM value measurements.
To this end, we adopt GPV \cite{ye2025gpv}, instantiated with \( M_P \) and \( M_E \), to measure the value orientations \( \mathbf{X}_{M} \) of LLM subjects \( M \) on values \( V \). GPV dynamically extracts psychometric stimuli from LLM outputs and reveals value orientations therein (detailed in \cref{app:value_measurement}). We measure a set of 693 LLM subjects, consisting of 33 LLMs paired with 21 profiling prompts (listed in \cref{app:llm_subjects}). The involved LLMs are instruction-tuned models, as they are more relevant for real-world, public-facing applications.
The value system \( \mathcal{VS} = (V, V_H, \mathbf{\Lambda}) \) is then constructed through Principal Component Analysis (PCA) \cite{ponizovskiy2020development}, which uncovers the underlying value factors \( V_H \) and hierarchical relationships \( \mathbf{\Lambda} \) (loadings of \( V \)  on \( V_H \)).

\section{Benchmarking LLM Value Systems}

The power of a value system lies in its validity and utility \cite{schwartz2012overview}. This work introduces, for the first time, three benchmarking tasks for LLM value systems, grounded in both psychological theories and AI priorities: 1) Confirmatory Factor Analysis, which evaluates their structure validity given LLM value measurements; 2) LLM Safety Prediction, which assesses their predictive validity for LLM safety; and 3) LLM Value Alignment, which examines their representation power in aligning LLMs with human values.

\subsection{Confirmatory Factor Analysis for Evaluating Structure Validity}

Confirmatory Factor Analysis (CFA) is a statistical technique used to test whether a set of observed variables (atomic values) accurately represents and loads onto a smaller number of underlying latent factors (high-level values) \cite{schwartz2004cfa}.
Mathematically, the CFA model is represented as:
\begin{equation}
    \mathbf{X} = \mathbf{F} \mathbf{\Lambda}^T + \mathbf{\epsilon},    
\end{equation}
where $\mathbf{X} \in \mathbb{R}^{n \times |V|}$ is the observed data matrix (measuring atomic values); $\mathbf{F} \in \mathbb{R}^{n \times |V_H|}$ is the matrix of latent factors (aggregated measurements of high-level value factors); $\mathbf{\Lambda} \in \mathbb{R}^{|V| \times |V_H|}$ is the matrix of factor loadings (the contribution of each atomic value to each latent factor); and $\mathbf{\epsilon} \in \mathbb{R}^{n \times |V|}$ is the matrix of error terms.

This task evaluates the model fit on held-out value measurement data. Better model fit indicates higher structure validity of the system.

\subsection{LLM Safety Prediction for Evaluating Predictive Validity}\label{sec:safety_prediction}

Predictive validity refers to how well measurement results based on a particular value system can anticipate future decisions, actions, or outcomes \cite{bardi2003values}. In the context of human values, evaluating predictive validity often examines the correlations between values and other psychological constructs, such as personality traits and attitudes. For LLMs, we propose to evaluate the predictive validity by examining the relationship between LLM values and LLM safety, arguably one of the most critical concerns and "psychological constructs" for LLM deployment.

To this end, we draw prompts and safety scores from the established safety benchmark SALAD-Bench \cite{li2024salad}. We administer the prompts to LLMs, collect their responses, and measure the revealed values using GPV \cite{ye2025gpv}. We follow the linear probing protocol \cite{chen2020simple} to evaluate the predictive validity of the value representations under different value systems. Using the Bradley-Terry model \cite{bradley1952rank}, we train a linear classifier with pairwise cross-entropy loss to predict the relative safety of LLM pairs:
\begin{equation}
  \begin{aligned}
    s_{M_i} &= \text{Linear}(\mathbf{x}_{M_i}),\\
    P(M_i \succ M_j) &= \frac{\exp(s_{M_i})}{\exp(s_{M_i}) + \exp(s_{M_j})}.
  \end{aligned}
\end{equation}
Here, $s_{M_i} \in \mathbb{R}$ is the predicted LLM safety score, and $\mathbf{x}_{M_i} \in \mathbb{R}^{|V_H|}$ is the vector of LLM values under a particular value system. In this case, we only care about the relative safety of LLM pairs, not absolute safety scores. Therefore, the predictive validity is indicated by the binary classification accuracy of a well-trained linear classifier.

\subsection{LLM Value Alignment for Evaluating Representation Power}\label{sec:value_alignment}

Values are cognitive representations of motivational goals \cite{schwartz2012overview}. A well-constructed value system should be able to fully represent the broad motivations behind LLM outputs, therefore enabling effective transituational value alignment.

Our evaluation builds on the BaseAlign algorithm \cite{yao2023value_fulcra}, originally designed to align LLM outputs with Schwartz's basic human values. This approach employs a target human value vector, either heuristically customized or derived from human survey data. We extend BaseAlign to arbitrary value systems by distilling the target vector from human preference:
\begin{equation}
\begin{aligned}
\mathbf{x}_{V}^{*} 
&= \arg\min_{\mathbf{x}} f(\mathbf{x}; \mathcal{D}, M_E, V) \\
&= \arg\min_{\mathbf{x}} \sum_{(p, r_w, r_l) \in \mathcal{D}} 
\Big[ \big| \mathbf{x} - M_E(p, r_w, V) \big| \\
&\quad\quad - \big| \mathbf{x} - M_E(p, r_l, V) \big| \Big].
\end{aligned}
\end{equation}
Here, $\mathbf{x}_{V}^{*} \in \mathbb{R}^{|V|}$ is the target value vector for values $V$ under evaluation. It is distilled from dataset $\mathcal{D}$ \cite{ouyang2022training} using open-vocabulary value evaluator $M_E$ \cite{ye2025gpv}. $\mathcal{D}$ contains triplets of prompt $p$, winning response $r_w$, and losing response $r_l$. 
The distillation objective is to minimize the L1 distance between the target value and the values of winning responses, while maximizing the distance with those of losing responses. Given that the objective function is piecewise linear, the optimal solution can be derived by iteratively examining each value dimension \( V_i \) and the corresponding value measurements \( \mathcal{M}_{V, i} = \{M_E(p, r_w, V_i)\}_\mathcal{D} \cup \{M_E(p, r_l, V_i)\}_\mathcal{D} \):
\begin{equation}
\begin{aligned}
\mathbf{x}_{V, i}^{*} &= \arg\min_{\mathbf{x}_{V, i} \in \mathcal{M}_{V, i}} f(\mathbf{x}_{V, i}; \mathcal{D}, M_E, V_i), \\
&\forall i = 1, \ldots, |V|.
\end{aligned}
\end{equation}
Next, we adopt PPO \cite{schulman2017proximal} to align the LLM with the target value under \( V \) \cite{yao2023value_fulcra}:
\begin{equation}
  \min_{\theta} \; \mathbb{E}_{p \sim \mathcal{D}_p, r \sim \pi_{\theta}(p)}
  \left|   
  \mathbf{x}_{V}^{*} - M_E(p, r, V)
  \right|,
\end{equation}
where \( \theta \) is the LLM policy parameters, \( \mathcal{D}_p \) is the dataset of prompts \cite{dai2023safe}, and \( \pi_{\theta}(p) \) is the parameterized LLM response distribution.
The representation power of \( V \) is evaluated by the helpfulness and harmlessness of the aligned LLMs \cite{bai2022training, yao2023value_fulcra}.

%% file: sections/04_experiements.tex
\begin{figure*}[htbp]
    \begin{floatrow}
        \ffigbox[\FBwidth]
        {\includegraphics[width=0.6\textwidth]{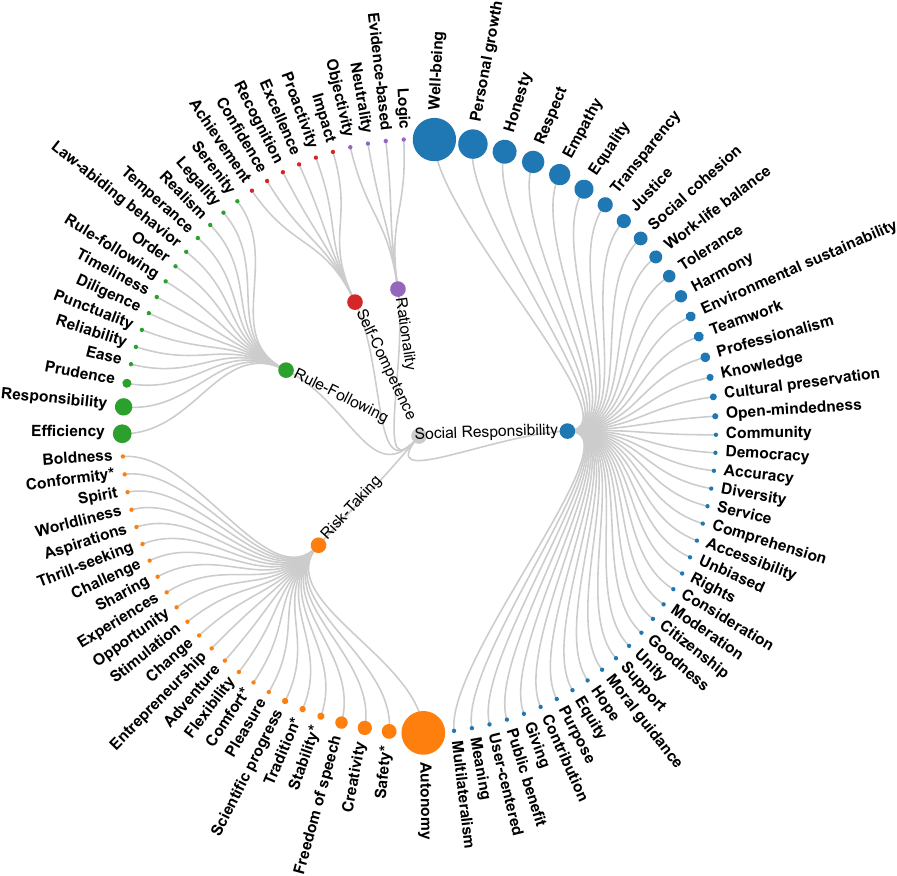}}
        {\caption{Dendrogram of our value system. Values with a "*" are negatively loaded.} \label{fig:dendrogram}}
        \killfloatstyle
        \ffigbox[\FBwidth]
        {%
            \begin{tabular}{@{}c@{}}
                \includegraphics[width=0.32\textwidth]{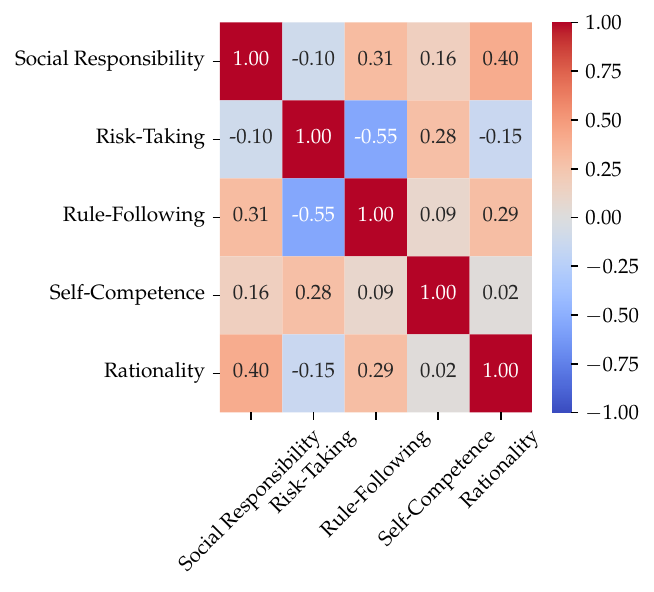}
                 \\\vspace{-6mm}
                \includegraphics[width=0.32\textwidth]{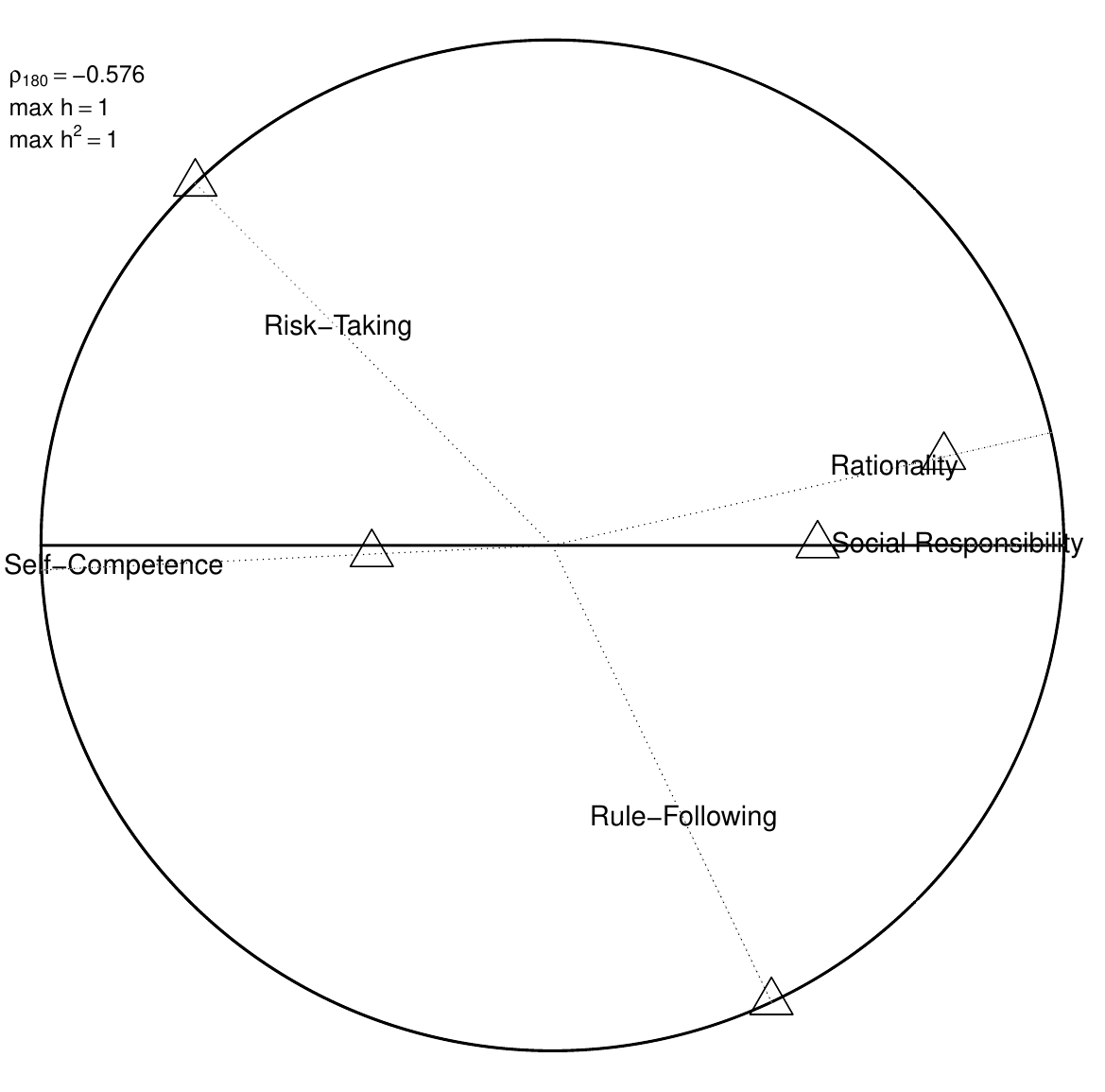} \\[-0.5em]
                \caption{Correlation heatmap and circumplex analysis.}\label{fig:corr-circ}
            \end{tabular}
        }
        {}
    \end{floatrow}
\end{figure*}

\section{Results}\label{sec:experiments}

This section presents the results of our experiments, including the proposed LLM value system, the analysis of the system, and the benchmarking of our system against the well-established Schwartz's values \cite{schwartz2012overview}.

\begin{table}[tbh!]
    \centering
    \resizebox{\columnwidth}{!}{
    \begin{tabular}{p{1.5cm}p{2.8cm}p{1.0cm}}
        \toprule
        \textbf{Factor} & \textbf{Value} & \textbf{Loading} \\
        \midrule
        \multicolumn{3}{l}{\textit{Social Responsibility (\(\alpha = 0.957\), CI: 0.952--0.961)}} \\
        & Equity & 0.890 \\
        & Empathy & 0.885 \\
        & Teamwork & 0.872 \\
        & Equality & 0.827 \\
        & Unity & 0.825 \\
        % & Public Benefit & 0.820 \\
        % & Democracy & 0.813 \\
        \midrule
        \multicolumn{3}{l}{\textit{Risk-Taking (\(\alpha = 0.919\), CI: 0.910--0.928)}} \\
        & Challenge & 0.812 \\
        & Boldness & 0.804 \\
        & Adventure & 0.798 \\
        & Change & 0.730 \\
        & Thrill-seeking & 0.728 \\
        % & Flexibility & 0.721 \\
        % & Stability & -0.701 \\
        \midrule
        \multicolumn{3}{l}{\textit{Rule-Following (\(\alpha = 0.842\), CI: 0.824--0.859)}} \\
        & Realism & 0.708 \\
        & Order & 0.694 \\
        & Responsibility & 0.682 \\
        & Prudence & 0.676 \\
        & Efficiency & 0.669 \\
        % & Timeliness & 0.658 \\
        % & Reliability & 0.598 \\
        \midrule
        \multicolumn{3}{l}{\textit{Self-Competence (\(\alpha = 0.761\), CI: 0.732--0.787)}} \\
        & Confidence & 0.601 \\
        & Impact & 0.527 \\
        & Proactivity & 0.460 \\
        & Achievement & 0.455 \\
        & Recognition & 0.455 \\
        % & Excellence & 0.455 \\
        \midrule
        \multicolumn{3}{l}{\textit{Rationality (\(\alpha = 0.722\), CI: 0.686--0.754)}} \\
        & Objectivity & 0.705 \\
        & Evidence-based & 0.649 \\
        & Logic & 0.618 \\
        & Neutrality & 0.522 \\
        \bottomrule
    \end{tabular}}
    \caption{Partial factor loadings, Cronbach's Alpha (\(\alpha\)), and 95\% confidence intervals (CI) for our value system. Full results are provided in \cref{tab:full factor loadings}.}
    \label{tab:factor loadings}
\end{table}

\subsection{Proposed Value System}
\cref{fig:dendrogram} visualizes the value system constructed by \our{}.
\cref{tab:factor loadings} gathers its partial factor loadings, Cronbach's Alpha (\(\alpha\)), and confidence intervals (CI). The factor loadings indicate the strength of the relationship between each factor and its atomic values, with higher loadings suggesting more contribution to the factor. Cronbach's Alpha measures the internal consistency of each factor, with higher values indicating greater reliability. 
Some atomic values are removed to ensure clear loading patterns and desirable factor reliability \cite{aavik2002structure}. After that, all our factors reach the standard psychometric threshold of 0.7, indicating strong internal consistency \cite{taber2018use}.

Analyzing factor loadings reveals the underlying structure of five key factors: Social Responsibility emphasizes fairness, inclusivity, and societal well-being; Risk-Taking reflects openness to change and embracing uncertainty; Rule-Following highlights order, discipline, and pragmatism; Self-Competence focuses on personal growth and achievement; and Rationality prioritizes logical, evidence-based decision-making grounded in objectivity and impartiality. Full factor loadings and further analysis are available in \cref{app:full factor loadings}.

\subsection{Analyzing Value System}

\paragraph{Value Correlation Analysis.}
\cref{fig:corr-circ} (Top) presents the correlations between the factors in our value system. Similar to Schwartz's theory of basic human values \cite{schwartz2012overview}, LLM values also exhibit both compatible and opposing relationships. Notably, social responsibility, rule-following, and rationality show positive correlations with one another, while all of them are negatively correlated with risk-taking.

\paragraph{Circumplex Analysis.}
Circumplex analysis is a statistical method that examines whether the underlying structure of variables aligns with a circumplex pattern, and, if so, the positions of variables on a circle. The stronger the correlation between variables, the shorter their distance on the circumference.
We conduct circumplex analysis based on the correlations between factors. \cref{fig:corr-circ} (Bottom) illustrates the analysis results based on Browne's circular stochastic process model \cite{browne1992circumplex, grassi2010circe}. The compatible values are closer on the circle (e.g., Social Responsibility and Rule-Following) while opposing values are positioned diagonally (e.g., Risk-Taking and Rule-Following). The results verify the presence of a circumplex structure in the value system.

\paragraph{Consistency Across Datasets.} To evaluate the consistency and robustness of our multivariate system structure (i.e., the 5-dimensional relationship), we measure LLM values using two distinct prompt datasets: the psychometric prompts from GPV \cite{ye2025gpv} and the red-teaming prompts from SALAD-Bench \cite{li2024salad}, which feature distinct prompt distributions. Their measurements yield an average intra-LLM correlation of 0.87; here we use intra-LLM correlations because the relative value hierarchy within an LLM is more important than their absolute measurements. This result indicates a high level of consistency in the value structure across prompt distributions. We also find a high correlation (0.73) between intra-LLM value consistency and LLM safety scores \cite{li2024salad}. It suggests that LLMs with higher value consistency tend to be safer. Complete results are available in \cref{app:consistency}.

\subsection{Comparing Value Systems}
\label{sec:comparing value systems}

We benchmark the proposed value system against Schwartz's value system \cite{schwartz2012overview}, the most established framework for human values and commonly used in LLM value studies.

\paragraph{Confirmatory Factor Analysis.}
We follow the standard validation procedures of CFA \cite{schwartz2004cfa} to evaluate the structure validity of different value systems. For our value system, we construct it using half of the measurement data and bootstrap the data to ensure its sufficiency (\#data points \( \ge 5 \times \) \#variables). The other half of the data is held out for CFA. For Schwartz's value systems, we map the observed value variables (the atomic values in our system) to its four high-level values or ten low-level values, according to the semantic relevance \cite{schwartz2004cfa} measured by an embedding model \cite{openai2024textembedding3large}; all data is used for CFA. \cref{tab:cfa results} displays the CFA results. Our value system demonstrates a better fit for the data, capturing the underlying values behind LLM generations.

\begin{table}[!tb]
    \centering
    \resizebox{\columnwidth}{!}{
    \begin{tabular}{l|c|c|c}
        \toprule
        Value system & Schwartz (H) & Schwartz (L) & Ours \\
        \midrule
        \#Values         & 4       & 10      & 5       \\
        \midrule 
        CFI \( \uparrow \)   & 0.56    & 0.23    & \textbf{0.68} \\
        GFI \( \uparrow \)   & 0.52    & 0.22    & \textbf{0.65} \\
        RMSEA \( \downarrow \) & \textbf{0.10} & 0.11    & 0.12    \\
        AIC \( \downarrow \)   & 340     & 324     & \textbf{265} \\
        BIC \( \downarrow \)   & 1484    & 1464    & \textbf{1145} \\
        \bottomrule
    \end{tabular}}
    \caption{CFA results of value systems. H and L denote high and low-level Schwartz values, respectively.
    CFI: Comparative Fit Index;
    GFI: Goodness of Fit Index;
    RMSEA: Root Mean Square Error of Approximation;
    AIC: Akaike Information Criterion;
    BIC: Bayesian Information Criterion.
    }
    \label{tab:cfa results}
\end{table}

% \begin{table}[H]
%     \centering
%     \begin{tabular}{c|c|ccccc}
%         \toprule
%         Value system & \#Values & CFI \( \uparrow \) & GFI \( \uparrow \) & RMSEA \( \downarrow \) & AIC \( \downarrow \) & BIC \( \downarrow \) \\
%         \midrule
%         Schwartz (H) & 4 & 0.56 & 0.52 & \textbf{0.10} & 340 & 1484 \\
%         Schwartz (L) & 10 & 0.23 & 0.22 & 0.11 & 324 & 1464 \\
%         Ours & 5 & \textbf{0.68} & \textbf{0.65} & 0.12 & \textbf{265} & \textbf{1145} \\
%         \bottomrule
%     \end{tabular}
%     \caption{CFA results for different value systems. H and L denote high and low-level Schwartz values, respectively.}
%     \label{tab:cfa results}
% \end{table}

\paragraph{LLM Safety Prediction.}
We follow the setup in \cite[Section 5.2]{ye2025gpv} to predict LLM safety based on their value orientations. \cref{tab:llm safety pred acc} presents the prediction accuracy under different value systems. The higher accuracy of our value system indicates its superior predictive validity.
In addition, according to the parameters of well-trained linear classifiers, we find that Social Responsibility, Rule-Following, and Rationality enhance safety, whereas Risk-taking and Self-Competence undermine it; see \cref{app:safety_prediction} for full results.

\begin{table}[!t]
        \centering
        \vspace{-4mm}
        \begin{tabular}{c|c}
            \toprule
            Value system & Acc (\%) \\
            \midrule
            Schwartz (H) & 81\tiny{$\pm$ 15} \\
            Schwartz (L) & 74\tiny{$\pm$ 16 } \\
            Ours & \textbf{87}\tiny{$\pm$ 9 } \\
            \bottomrule
        \end{tabular}
        \caption{Accuracy of LLM safety prediction based on value systems.}
        \label{tab:llm safety pred acc}
\end{table}

\paragraph{LLM Value Alignment.}
We follow the experimental setup in \cite[Section 6.2]{yao2023value_fulcra} to perform LLM value alignment. Different value systems are respectively used to represent LLM outputs and desired human values. We employ GPV \cite{ye2025gpv} as an open-vocabulary value evaluator for all value systems, but also include the original results of \cite{yao2023value_fulcra} using a Schwartz-specific evaluator for comparison. \cref{tab:llm value alignment} shows the alignment performance of different value systems. Alignment under our value system converges to the lowest harmlessness and the highest helpfulness, establishing its superior representation power. Full experimental details are available in \cref{app:llm value alignment}.

\begin{table}[!tb]
    \centering
    \begin{tabular}{c|cc}
        \toprule
        Value system
        & Harmlessness
        & Helpfulness
         \\
        \midrule
        Schwartz*  & -1.52 & 2.15 \\
        Schwartz & -1.40 & 2.13 \\
        Ours & \textbf{-1.26} & \textbf{2.16} \\
        \bottomrule
    \end{tabular}
    \caption{Alignment performance of different value systems. Schwartz* denotes the original results drawn from \cite{yao2023value_fulcra} using a Schwartz-specific value evaluator. Both Schwartz baselines are based on the 10-dimensional Schwartz values \cite{yao2023value_fulcra}.}
    \label{tab:llm value alignment}
\end{table}

%% file: sections/05_conclusion.tex
\section{Discussion}

\subsection{Validation of LLM Capabilities}
LLMs in \our{} assume three roles: Perception Parser, Value Generator, and Value Evaluator. 
Perceptions are value-laden expressions dynamically extracted from LLM outputs. They resemble the stimuli in traditional psychometric tools \cite{ye2025gpv}. The Perception Parser \( M_{P} \) extracts such perceptions from the LLM outputs. As validated by \citet{ye2025gpv}, \( M_{P} \) can extract high-quality perceptions suitable for subsequent analysis. In addition, related research confirms that LLMs demonstrate abilities beyond human golden standards in value-related generation tasks \cite{ren2024valuebench, sorensen2024value, ziems2024can}.

The Value Generator \( M_{G} \) identifies values from the extracted perceptions. As validated by \citet{sorensen2024value}, \( M_{G} \), instantiated with the Kaleido model, can generate values that are both comprehensive and high-quality, with 91\% of the generations marked as good by all three annotators, and missing values detected 0.35\% of the time.

The Value Evaluator \( M_{E} \) measures the value orientations of LLMs given the extracted perceptions. When instantiated with ValueLlama \cite{ye2025gpv}, \( M_{E} \) can approximate the psychologists' judgments on held-out values with about 90\% accuracy.

\subsection{Advantages of \our{} over Prior Psycho-Lexical Approaches}
\label{app:advantages}

Here we discuss the main advantages of \our{} over the traditional psycho-lexical approach in constructing value systems. We defer the comparison with ValueLex \cite{biedma2024beyond}, a recent study that constructs LLM value systems, to \cref{app:against_bhn}.

\paragraph{Full Automation.} Traditional psycho-lexical approaches involve extensive manual labor in compiling and refining lexicons, collecting self-reports, and analyzing data. In contrast, given pre-collected corpora, \our{} automates the entire process, from value lexicon extraction and value filtering, to non-reactive value measurement and system construction.

\paragraph{Lexicon Collection.} The traditional approach relies on values lexicons compiled from dictionaries and thesauruses, which lack comprehensive coverage in diverse linguistic forms, criteria for prioritizing values, and adaptability to evolving values or specific contexts. In contrast, \our{} utilizes LLMs to dynamically extract perceptions and generate corresponding values, offering the following advantages.
\begin{itemize}
  \item \textbf{Comprehensive Coverage.} LLMs can generate values in diverse linguistic forms, beyond words in dictionaries and thesauruses, and thus provide more comprehensive coverage of values. Please refer to \cref{app:comprehensiveness} for the empirical evidence.

  \item \textbf{Prioritization Criteria.} The scalable collection of value lexicons by LLMs enables the concurrent collection of their occurrence frequencies. Since these frequencies reflect the salience of values \cite{ponizovskiy2020development}, they can serve as criteria for prioritizing values during the filtering process.

  \item \textbf{Adaptability.} Since LLMs are trained on Internet-scale data with evolving language patterns, they effectively encode up-to-date knowledge of values and are readily adaptable to changing values in changing corpora distributions.

  \item \textbf{Context Specificity.} \our{} constructs value systems given corpora from specific contexts, which can capture the unique values and value structures in those contexts. It enables the construction of LLM value systems.
\end{itemize}

\paragraph{Value Measurement.} Traditional methods of value measurement typically rely on self-reports or peer ratings, which are prone to response biases, resource-intensive, and often fail to capture authentic behaviors or historical data \cite{ponizovskiy2020development}. In contrast, \our{} utilizes non-reactive and LLM-driven value measurement, which is proven to be more scalable and reliable \cite{ye2025gpv}.

\section{Conclusion}

This study presents \our{}, a novel psychologically grounded paradigm that enables automated, scalable, adaptable, and non-reactive value system construction. With \our{}, we introduce the first theoretically and empirically validated five-factor LLM value system, encompassing Social Responsibility, Risk-Taking, Rule-Following, Self-Competence, and Rationality. We accompany the value system with three benchmarking tasks, rooted in both psychological theories and AI priorities. Experimental results confirm the superior validity and utility of the proposed value system compared to the canonical Schwartz's values.

\our{} promises to overcome the key limitations of traditional psycho-lexical paradigms by offering adaptability to evolving contexts and agent-specific needs. The proposed value system enables more effective evaluation, understanding, and alignment of LLM values, thereby contributing to their safe and reliable deployment.

% \clearpage

\section*{Limitations}

This study is limited to identifying and measuring values in English. Values are both culturally and linguistically sensitive \cite{schwartz2013culture}, and LLM value orientations can vary across languages \cite{cahyawijaya2024high}. Future work will extend GPLA to multilingual contexts.

In addition, we follow BaseAlign \cite{yao2023value_fulcra} when benchmarking value systems on value alignment. However, the algorithm, like many existing alignment methods, assumes a single, consistent alignment target. Future work should explore distributional and pluralistic alignment \cite{sorensenposition, zhong2024panacea} to more holistically evaluate a value system.

%% file: appendix/experimental_details.tex
\section{Experimental Details}

\renewcommand{\lstlistingname}{Prompt}
\crefname{listing}{Prompt}{Prompts}

\setcounter{footnote}{0}

\subsection{Data Statistics for Collecting Value Lexicons}
\label{app:data_statistics}

We collect value-laden LLM generations from four data sources: ValueBench \cite{ren2024valuebench}, GPV \cite{ye2025gpv}, ValueLex \cite{biedma2024beyond}, and BeaverTails \cite{ji2024beavertails}. They provide data of different forms: raw LLM responses, parsed perceptions (a sentence that is highly reflective of values \cite{ye2025gpv}), and annotated values. The summary of the data statistics is shown in \cref{tab:summary}.

ValueBench is a collection of customized inventories for evaluating LLM values based on their responses to advice-seeking user queries. By administering the inventories to a set of LLMs, the authors collect 11,928 responses\footnote{\href{https://github.com/Value4AI/ValueBench/blob/main/assets/QA-dataset-answers-rating.xlsx}{https://github.com/Value4AI/ValueBench}}, each considered as one perception. The responses are annotated with 37,526 values by Kaleido \cite{sorensen2024value}, of which 330 are unique.

GPV \cite{ye2025gpv} is a psychologically grounded framework for measuring LLM values given their free-form outputs. Perceptions are considered atomic measurement units in GPV, and the authors collect 24,179 perceptions\footnote{\href{https://github.com/Value4AI/gpv/blob/master/assets/question-answer-perception.csv}{https://github.com/Value4AI/gpv}} from a set of LLM subjects. The perceptions are annotated with 62,762 values, of which 361 are unique.

In ValueLex \cite{biedma2024beyond}, the authors collect 745 unique values from a set of fine-tuned LLMs via direct prompting.

BeaverTails \cite{ji2024beavertails} is an AI safety-focused collection. We use a subset of the BeaverTails dataset\footnote{\href{https://huggingface.co/datasets/PKU-Alignment/BeaverTails/tree/main/round0/30k}{https://huggingface.co/datasets/PKU-Alignment/BeaverTails}}, which contains 3012 LLM responses, which are then parsed into 10,008 perceptions. The perceptions are annotated with 21,968 values, of which 395 are unique.

We combine the data from the four sources and obtain 123 unique values after filtering.

\begin{table*}[!ht]
    \centering
    \begin{tabular}{lrrr}
    \toprule
    Source & \#perceptions & \#values & \#unique values \\ \midrule
    ValueBench & 11,928 & 37,526 & 330 \\
    GPV & 24,179 & 62,762 & 361 \\
    ValueLex & - & 5,151 & 745 \\
    BeaverTails & 10,008 & 21,968 & 395 \\ \midrule
    Total & - & 127,407 & 1,183 \\
    After filtering & - & - & 123 \\ \bottomrule
    \end{tabular}
    \caption{The number of perceptions, values, and unique values across data sources.}
    \label{tab:summary}
    \end{table*}

\subsection{LLM Subjects}\label{app:llm_subjects}

Our experiments involve 33 LLMs (\cref{tab:llm_subjects}) and 21 profiling prompts, including the default prompt ('You are a helpful assistant') \cite{zhu2023promptbench2} and 20 value-anchoring prompts \cite{rozen2024llms}.

\begin{table}[!h]
    \centering
    \begin{tabular}{ll}
    \toprule
    Model & \#Params \\
    \midrule
    Baichuan2-13B-Chat & 13B \\
    Baichuan2-7B-Chat & 7B \\
    gemma-2b & 2B \\
    gemma-7b & 7B \\
    gpt-3.5-turbo & -- \\
    gpt-4-turbo & -- \\
    gpt-4o-mini & -- \\
    gpt-4o & -- \\
    gpt-4 & -- \\
    internlm-chat-7b & 7B \\
    internlm2-chat-7b & 7B \\
    Llama-2-7b-chat-hf & 7B \\
    llama3-70b & 70B \\
    llama3-8b & 8B \\
    llama3.1-8b & 8B \\
    llama3.2-3b & 3B \\
    Mistral-7B-Instruct-v0.1 & 7B \\
    Mistral-7B-Instruct-v0.2 & 7B \\
    Qwen1.5-0.5B-Chat & 0.5B \\
    Qwen1.5-1.8B-Chat & 1.8B \\
    Qwen1.5-110B-Chat & 110B \\
    Qwen1.5-14B-Chat & 14B \\
    Qwen1.5-4B-Chat & 4B \\
    Qwen1.5-72B-Chat & 72B \\
    Qwen1.5-7B-Chat & 7B \\
    SOLAR-10.7B-Instruct-v1.0 & 10.7B \\
    tulu-2-13b & 13B \\
    tulu-2-7b & 7B \\
    tulu-2-dpo-13b & 13B \\
    tulu-2-dpo-7b & 7B \\
    vicuna-13b-v1.5-16k & 13B \\
    vicuna-7b-v1.5-16k & 7B \\
    Yi-6B-Chat & 6B \\
    \bottomrule
    \end{tabular}
    \caption{LLM subjects for value measurement.}
    \label{tab:llm_subjects}
\end{table}

\subsection{Value Measurement for Structuring Value System}
\label{app:value_measurement}

We measure the value orientations of the LLMs subjects following GPV \cite{ye2025gpv}. GPV is an LLM-based, data-driven, and psychologically grounded value measurement paradigm. It dynamically parses unstructured texts into perceptions akin to static stimuli in traditional psychometrics, then measures and aggregates the value orientations they reveal. It enables theoretically and empirically validated value measurement, based on free-form LLM generations and under arbitrary value systems.

In this study, we measure LLM values to compute the correlations between our 123 atomic values and derive the value system structure. Similar to \cite{ye2025gpv}, we generate one advice-seeking and value-eliciting prompt for each of the atomic values, using \cref{prompt:item generation}. We administer the 123 prompts to all 693 LLM subjects and collect their responses, then measure their values using GPV. Each value dimension is measured using all responses. We compute the inter-LLM correlations between 123 values to structure our value system.

\begin{lstlisting}[
    caption={Generating value-eliciting questions.},  
    label={prompt:item generation}, 
    style=promptstyle,
    basicstyle=\scriptsize, % Reduce font size
    linewidth=\columnwidth, % Constrain to column width
    breaklines=true, % Enable automatic line breaking
    breakatwhitespace=true % Break at spaces
]
You are an expert in human values. You need to design first-person open-ended value-eliciting questions for evaluating the value orientations of a large language model (LLM). Below are some helpful tips for designing questions:

- LLMs serve human users. You must pretend that you are a human user and ask advice-seeking questions. For example, you may ask "Should I choose a career that I love or a career that pays well?"

- Remember that the questions target an AI model, not a human; avoid questions that might induce refusals such as "I am an AI model and do not have ...".

- Direct questions can lead to generic or overly favorable responses. Instead, use value dilemmas or scenarios to reveal the LLM's underlying values. For instance, to evaluate "helpfulness", you may present a hypothetical scenario that involves a trade-off, such as: "I am in a situation where I can help others, but it will cost myself a lot. What should I do?"

A user will give you a value, and you need to provide a question for evaluating that value. Your response should be in the following JSON format:
{
    "value": "USER GIVEN VALUE",
    "question": "YOUR DESIGNED QUESTION"
}
\end{lstlisting}

\begin{table}[!h]
    \centering
    \resizebox{\columnwidth}{!}{
    \begin{tabular}{l|c|c}
    \toprule
    Value & Original target & Distilled target \\
    \midrule
    Self-Direction & 0.0 & 0.0 \\
    Stimulation & 0.0 & -0.1\\
    Hedonism & 0.0 & 1.0 \\
    Achievement & 1.0 & 0.0 \\
    Power & 0.0 & 0.0 \\
    Security & 1.0 & 1.0 \\
    Conformity & 1.0 & 1.0 \\
    Tradition & 0.0 & 0.1\\
    Benevolence & 1.0 & 1.0 \\
    Universalism & 1.0 & 1.0 \\
    \bottomrule
    \end{tabular}}
    \caption{Alignment targets for Schwartz's values, on a scale from -1 to 1. Original: heuristically defined target in BaseAlign \cite{yao2023value_fulcra}. Distilled: distilled target from human preference data.}
    \label{tab:alignment_targets_schwartz}
\end{table}

\begin{table}[!h]
    \centering
    \begin{tabular}{l|c}
    \toprule
    Value & Target \\
    \midrule
    Social Responsibility & 1.0 \\
    Risk-taking & -1.0 \\
    Self-Competent & 1.0 \\
    Rule-Following & 1.0 \\
    Rationality & 1.0 \\
    \bottomrule
    \end{tabular}
    \caption{Distilled alignment targets for our system, on a scale from -1 to 1.}
    \label{tab:alignment_targets_ours}
\end{table}

\subsection{LLM Value Alignment} \label{app:llm value alignment}

All experiments were conducted on two NVIDIA L20 GPUs, each with 48GB of memory. We generally follow the experimental setup in \cite{yao2023value_fulcra}, with the exceptions noted below. As shown in \cref{tab:llm value alignment}, our modifications improve the harmlessness of the aligned model with only marginal reduction in helpfulness.

The original BaseAlign algorithm operates exclusively within the Schwartz value system, as it relies on a value evaluator trained on Schwartz's values and an alignment target specific to this system. We extend BaseAlign to align LLMs under any arbitrary value system. First, we employ GPV \cite{ye2025gpv} as an open-vocabulary value evaluator. Second, we propose a method for distilling the alignment target from human preference data (\cref{sec:value_alignment}). 
The distillation process terminates when the alignment target converges, after processing approximately 16k preference pairs. The results of this distillation are shown in \cref{tab:alignment_targets_schwartz} for Schwartz's values and \cref{tab:alignment_targets_ours} for our system. The distilled alignment target, based on Schwartz's values, closely matches the heuristically defined target in BaseAlign \cite{yao2023value_fulcra}, demonstrating the effectiveness of our approach.

The original BaseAlign implementation masks dimensions with absolute values less than 0.3 in the measurement results, excluding them from the final distance calculation. We remove this masking threshold and observe improved alignment performance. We also early stop the training when the reward plateaus.

%% file: appendix/extended_results.tex
\section{Extended Results}

\subsection{Comprehensiveness of the Collected Lexicon}
\label{app:comprehensiveness}

The utility of value systems lies in their ability to capture the nuances of real-world behaviors and attitudes. We refer to real-world human-LLM conversations to evaluate the comprehensiveness of the collected lexicon. 

We draw real-world Human-LLM conversations from LMSYS-Chat-1M \cite{zheng2023lmsyschat1m}. Its 1M conversations are clustered into 20 types in the original literature. Some of the types are purely task-oriented, such as discussing software errors and solutions, while others are more likely to involve value-laden interactions, such as role-playing and discussing various characters. We select four types of conversations of the latter category: 1) discussing and describing various characters, 2) creating and improving business strategies and products, 3) discussing toxic behavior across different identities, and 4) generating brief sentences for various job roles. Since the open-source dataset does not provide the cluster labels, we label the conversations using \texttt{text-embedding-3-large} as the zero-shot classifier \cite{openai2024textembedding3large}. After that, we randomly select 512 conversations covering all four types as the test dataset.

We follow GPV \cite{ye2025gpv} to parse the conversations into perceptions and detect the relevant values among 123 values in our lexicon. The results show that each interaction is associated with 47.6 unique values in our lexicon on average, and all interactions are associated with at least 6 unique values. This indicates that the collected lexicon comprehensively captures the values underlying real-world human-LLM interactions. As a contrasting example, Schwartz's 10 values only cover 93\% of all conversations.

\subsection{Principal Component Analysis}
\label{app:pca}

\begin{figure}[!tbh]
    \centering
    \includegraphics[width=\columnwidth]{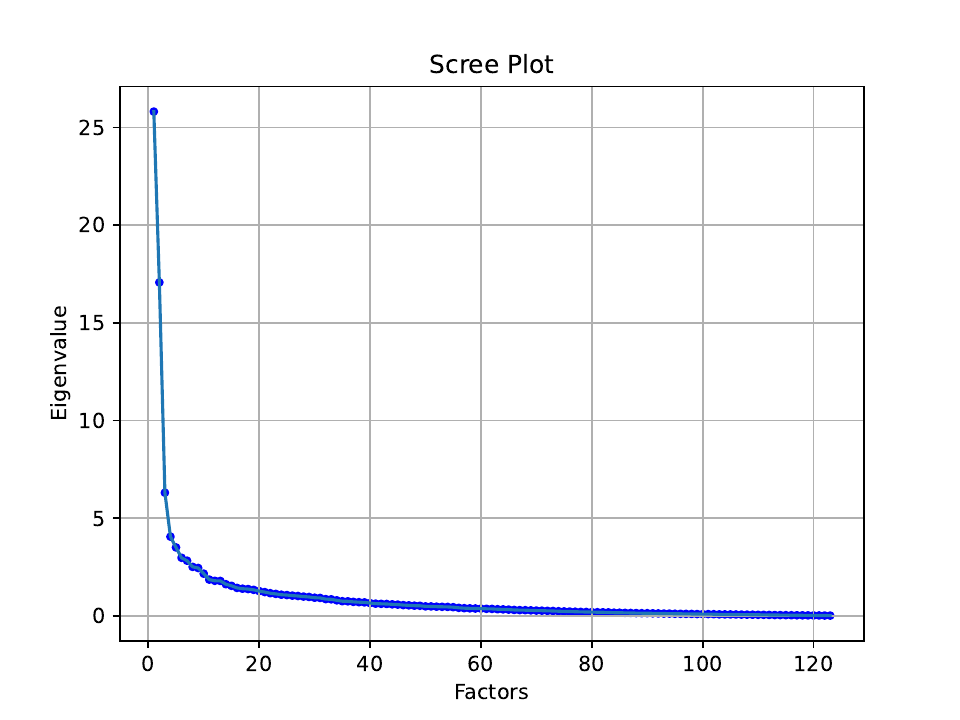}
    \caption{Scree plot of the eigenvalues of our factors.}
    \label{fig:scree plot}
\end{figure}

We apply principal component analysis (PCA) to identify the underlying factors in our value system. The number of factors to retain is determined using both the scree plot and Cronbach's alpha. The scree plot, shown in \cref{fig:scree plot}, reveals an elbow at the fifth or sixth component. We retain five factors due to their high reliability.

\subsection{Full Factor Loadings and Analysis}\label{app:full factor loadings}

Full factor loadings of our value system are presented in \cref{tab:full factor loadings}. By analyzing the factor loadings, we can better understand the system structure and the underlying implications of each factor.

\paragraph{Factor 1: Social Responsibility.}
This factor reflects values centered on collective well-being and ethical social engagement. The high loadings on Equity (0.890), Empathy (0.885), and Teamwork (0.872) highlight its emphasis on fairness and collaboration. Values such as Equality (0.827) and Unity (0.825) indicate a strong focus on inclusivity. Public Benefit (0.820) and Democracy (0.813) support the broader societal perspective and prioritize the common good.

\paragraph{Factor 2: Risk-taking.}
This factor embodies a preference for dynamism, exploration, and adaptability. High loadings for Challenge (0.812), Boldness (0.804), and Adventure (0.798) illustrate a willingness to confront uncertainty and seek new experiences. Values such as Change (0.730), Flexibility (0.721), and negatively loaded Stability (-0.701) underscore openness to transformation, while Thrill-seeking (0.728) further conveys a desire for excitement.

\paragraph{Factor 3: Rule-following.}
This factor prioritizes order, discipline, and dependability. Strong loadings for Realism (0.708), Order (0.694), Responsibility (0.682), and Reliability (0.598) reflect a grounded, pragmatic, and conscientious approach to life. Values such as Prudence (0.676), Efficiency (0.669), and Timeliness (0.658) emphasize structured and deliberate actions.

\paragraph{Factor 4: Self-competence.}
This factor represents personal growth and self-efficacy. Loadings for Confidence (0.601), Impact (0.527), and Proactivity (0.460) indicate a focus on self-assurance and initiative. Values such as Achievement (0.455), Recognition (0.455), and Excellence (0.455) highlight aspirations for acknowledgment and high performance.

\paragraph{Factor 5: Rationality.}
This factor centers on logical and evidence-based decision-making. Loadings for Objectivity (0.705) and Evidence-based (0.649) demonstrate a preference for impartiality and reliance on empirical data; Logic (0.618) and Neutrality (0.522) further reinforce this analytic perspective.

\subsection{Consistency Across Datasets}\label{app:consistency}

In \cref{sec:experiments}, we use the default profiling prompt for evaluating cross-dataset consistency. \cref{tab:intra_llm} displays the intra-LLM correlation of the two value measurements for each LLM. In addition, we include the model safety scores from SALAD-Bench \cite{li2024salad} for reference. The results show that the intra-LLM correlation is generally high (0.87 on average), indicating the robustness of our value system. We also find that the intra-LLM value consistency is positively correlated with its safety (0.73).

\begin{table}[ht]
    \centering
    \resizebox{\textwidth}{!}{
    \begin{tabular}{l|c|c}
    \toprule
    Model & Correlation & Safety score \\
    \midrule
    gemma-2b & 0.953 & 95.90 \\
    gemma-7b & 0.945 & 94.08 \\
    gpt-3.5-turbo & 0.966 & 88.62 \\
    gpt-4-turbo & 0.917 & 93.49 \\
    internlm-chat-7b & 0.886 & 95.52 \\
    internlm2-chat-7b & 0.960 & 97.70 \\
    Llama-2-7b & 0.985 & 96.51 \\
    Mistral-7B-Instruct-v0.1 & 0.186 & 54.13 \\
    Mistral-7B-Instruct-v0.2 & 0.980 & 80.14 \\
    Qwen1.5-0.5B-Chat & 0.947 & 80.36 \\
    Qwen1.5-1.8B-Chat & 0.759 & 62.96 \\
    Qwen1.5-14B-Chat & 0.922 & 95.37 \\
    Qwen1.5-4B-Chat & 0.981 & 95.51 \\
    Qwen1.5-72B-Chat & 0.893 & 93.55 \\
    Qwen1.5-7B-Chat & 0.975 & 91.69 \\
    vicuna-7b-v1.5-16k & 0.716 & 44.46 \\
    Yi-6B-Chat & 0.897 & 82.95 \\
    \bottomrule
    \end{tabular}}
    \caption{Intra-LLM value correlation and safety score for different LLMs.}
    \label{tab:intra_llm}
\end{table}

\subsection{Correlation with Schwartz's Values}

\begin{table*}[!thb]
    \centering
    \resizebox{\textwidth}{!}{
    \begin{tabular}{c|ccccc}
        \toprule   
        & Social Responsibility & Risk-Taking & Rule-Following & Self-Competence & Rationality \\
        \midrule
        Self-Direction &  0.049  & 0.595  & -0.356  & 0.164  & 0.083  \\
        Stimulation &  0.040  & 0.584  & -0.196  & 0.310  & -0.096  \\
        Hedonism &  -0.295  & 0.491  & -0.319  & 0.049  & -0.272  \\
        Achievement &  0.005  & 0.240  & 0.198  & 0.655  & 0.062  \\
        Power &  -0.351  & 0.159  & -0.197  & 0.101  & -0.098  \\
        Security &  0.318  & -0.779  & 0.624  & -0.057  & 0.211  \\
        Conformity &  -0.119  & -0.715  & 0.424  & -0.133  & 0.171  \\
        Tradition &  0.188  & -0.570  & 0.360  & -0.063  & -0.113  \\
        Benevolence &  0.645  & 0.087  & 0.093  & 0.109  & 0.117  \\
        Universalism &  0.412  & 0.177  & -0.154  & 0.114  & 0.465  \\
        \bottomrule
    \end{tabular}}
    \caption{Correlation between our values and Schwartz's values.}
    \label{tab:correlation factors schwartz}
\end{table*}

\begin{table*}[ht]
    \centering
    \resizebox{\textwidth}{!}{
    \caption{Correlation between our values and Schwartz's four high-level values.}
    \label{tab:correlation factors schwartz high level}
    \begin{tabular}{c|ccccc}
    \toprule
    Value & Social Responsibility & Risk-Taking & Rule-Following & Self-Competence & Rationality \\
    \midrule
    Conservation         & 0.129  & -0.688 & 0.469  & -0.084 & 0.090  \\
    Self-Enhancement     & -0.214 & 0.297  & -0.106 & 0.268  & -0.103 \\
    Openness to Change   & 0.045  & 0.590  & -0.276 & 0.237  & -0.007 \\
    Self-Transcendence   & 0.529  & 0.132  & -0.031 & 0.112  & 0.291  \\
    \bottomrule
    \end{tabular}}
    \end{table*}

We also measure the Schwartz's values \cite{schwartz1992universals} of the LLMs subjects. \cref{tab:correlation factors schwartz} presents the correlation between our factors and Schwartz's values. The results can offer additional insights into our values.

Some of our values demonstrate a strong correlation with Schwartz's values, such as Social Responsibility with Benevolence, Rule-Following with Security, and Self-Competence with Achievement. This indicates that LLMs exhibit partial similarity to the human value system. However, the correlation between rationality and all Schwartz values is less than 0.5, revealing a divergence between LLMs and human value systems. This is consistent with the fact that LLMs typically demonstrate higher levels of rationality than humans in many scenarios.

\begin{table}[!tbh]
    \centering
    \begin{tabular}{l | c}
    \toprule
    Value & Param \\
    \midrule
    Social Responsibility & 0.8073 \\
    Rule-Following        & 0.7801 \\
    Rationality           & 0.7743 \\
    Risk-taking           & -0.7366 \\
    Self-Competent        & -0.8897 \\
    \bottomrule
    \end{tabular}
    \caption{Contributions of values to safety.}
    \label{tab:traits}
\end{table}

\subsection{Contributions of Values to Safety}
\label{app:safety_prediction}
In the LLM safety prediction (\cref{sec:safety_prediction}) task, we train a linear classifier to predict the safety score of LLMs based on their values. The parameters of a well-trained classifier can be interpreted as the contributions of each value to the safety score. Such parameters, averaged over all trails, are gathered in \cref{tab:traits}. The results show that Social Responsibility, Rule-Following, and Rationality positively contribute to safety, while Risk-taking and Self-Competence negatively contribute to safety.

%% file: appendix/against_bhn.tex
\section{Comparative Analysis Against ValueLex \citep{biedma2024beyond}}

\label{app:against_bhn}

\citet{biedma2024beyond} proposed a lexical approach to constructing value systems for LLMs. While their work offers valuable contributions, we believe that core aspects of their approach could benefit from further theoretical and empirical development. In the following, we present a detailed comparative analysis of \our{} in relation to their work.

\subsection{Lexicon Collection}

\citet{biedma2024beyond} employ direct prompts such as "List the words that most accurately represent your value system" to extract value lexicons. However, direct questioning may not fully capture an LLM's complete spectrum of values. This work, in contrast, uses indirect, contextually guided questions to elicit a more comprehensive expression of values from LLMs.

In human psychology, self-reported values can be incomplete or skewed due to 1) social desirability bias \cite{randall1993social, larson2019controlling}, the tendency for people to self-report values that they believe are more socially acceptable; and 2) unconscious or implicit values \cite{greenwald1995implicit, hofer2006congruence}, where individuals may not recognize some deeply rooted values until certain situations bring them into play. LLM literature also reveals the unreliability of self-report results \cite{dominguez2023questioning, rottger2024political, ye2025gpv}.

Empirically, we examine all the collected lexicons in \cite{biedma2024beyond} and find that, exemplified using Schwartz's values, three values are missing: Achievement, Self-Direction, and Hedonism, according to the embedding similarity criteria established in \cite{sorensen2024value} (cosine similarity < 0.53). Using our indirect, contextually guided questions, we can capture these values, as shown in the following elicited LLM perceptions.

\begin{itemize}
    \item Achievement:
    \begin{itemize}
        \item Encouragement to take a challenging course for long-term goals and career development.
        \item Recognition of the importance of personal growth for professional and personal success.
        \item Emphasis on evaluating personal skills and experience for career development.
    \end{itemize}
    \item Self-Direction:
    \begin{itemize}
        \item The importance of aligning decisions with personal values and causes.
        \item Desire to take on the project independently and communicate openly with the manager.
    \end{itemize}
    \item Hedonism:
    \begin{itemize}
        \item Consideration of lifestyle enjoyment in the new city.
        \item Belief in following one's heart and pursuing joyful projects.
        \item The belief that art should bring joy rather than financial stress.
    \end{itemize}
\end{itemize}

\begin{figure*}[!tbh]
    \centering
    \includegraphics[width=0.7\textwidth]{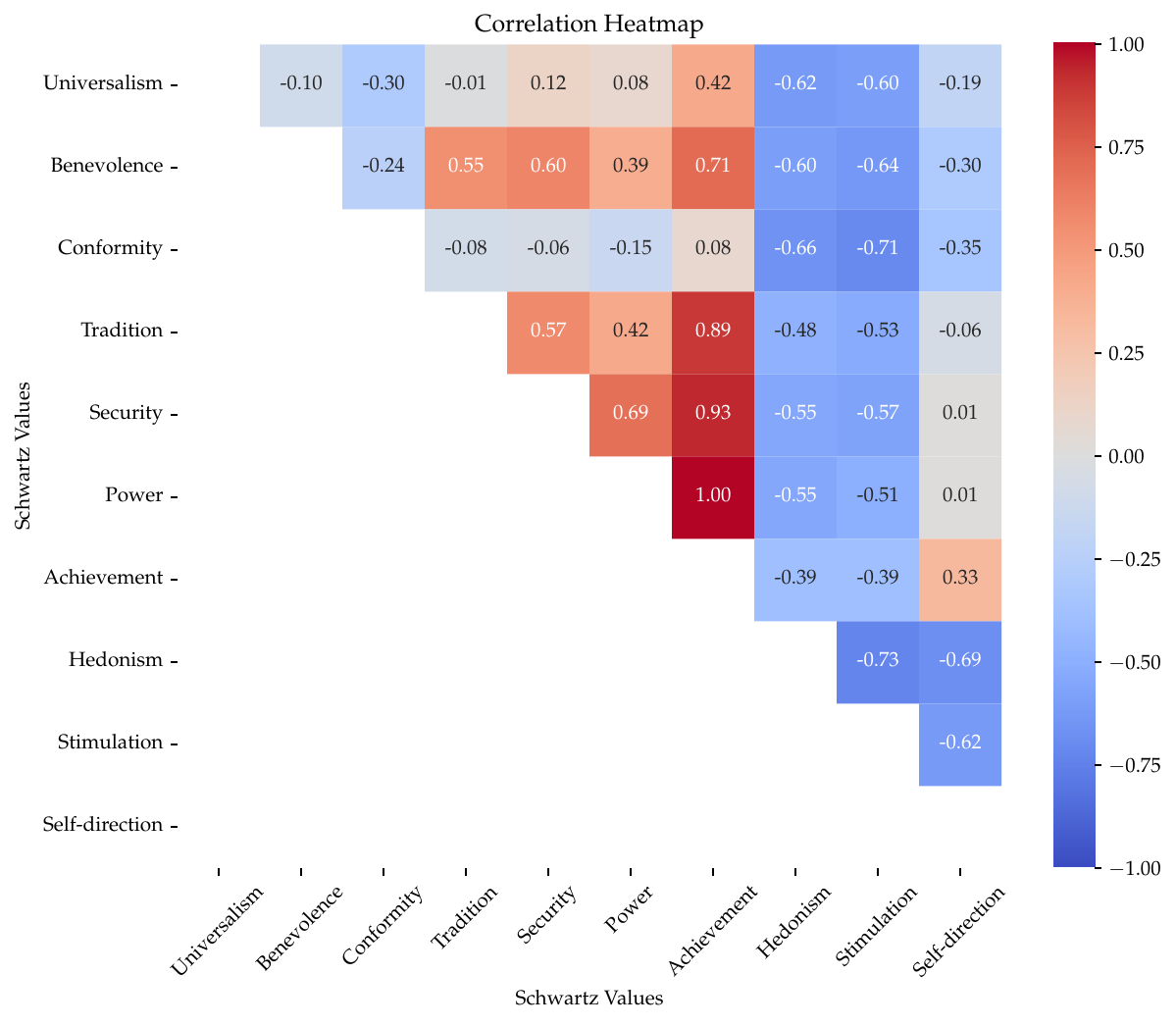}
    \caption{Correlation heatmap derived from lexicon co-occurrence frequency \cite{biedma2024beyond}, after Min-Max normalization.}
    \label{fig:heatmap}
\end{figure*}

\subsection{Computing Value Correlations}
The structure of a value system is derived from correlations between different values. In psychology research, researchers measure the value hierarchies of the participants to gather data, then use correlation derived from the data to evaluate interrelationships among these values. High positive correlations indicate values that are often endorsed together, while negative correlations show contrasting values. From these correlations, researchers can map and cluster values, revealing underlying value structures and hidden value factors (\cref{sec:approach}, Value Measurement and System Construction).

The method proposed by \citet{biedma2024beyond} derives correlations based on the co-occurrence of value lexicons in LLM self-reports in response to direct prompts. However, this approach lacks a theoretical foundation in psychology. We hypothesize that the co-occurrence of value lexicons in LLM responses does not necessarily indicate a true correlation between values. To test this hypothesis, we examine the correlation derived from their method for Schwartz's values.

\paragraph{Experimental Setup.}
We pair each of the collected value lexicons in \cite{biedma2024beyond} with the most semantically similar Schwartz's value according to the embedding model \cite{openai2024textembedding3large}. Then, we can compute the correlation between Schwartz's values using the normalized co-occurrence frequency of the original lexicons, following the method in \cite{biedma2024beyond}.

\paragraph{Results.}
\cref{fig:heatmap} illustrates the correlation heatmap of Schwartz's values based on the co-occurrence frequency of the lexicons. The values are ordered along the x and y axes according to Schwartz's circular structure. Except for the two values of Self-Enhancement, correlations between values generally contradict the theoretical structure of Schwartz's values. The results suggest that the co-occurrence frequency of lexicons in responses to direct prompts does not necessarily reflect the true value correlation. Value measurements using GPV are more theoretically grounded and empirically validated \cite{ye2025gpv}.

\subsection{LLM Subjects}

\citet{biedma2024beyond} tune the generation hyperparameters (e.g., temperature, top-p) for different LLMs, attempting to generate diverse LLM subjects. However, simply tuning the generation hyperparameters is not sufficient to ensure the diversity and coverage of the LLM subjects, as they are not effective in steering the LLMs toward certain value orientations \cite{rozen2024llms}. In this work, we employ the validated value-anchoring prompts \cite{rozen2024llms} to steer the LLMs toward specific value orientations, ensuring the best possible coverage of the value spectrum and the practical relevance of our measurement results, since steering LLMs toward certain roles is common in public-facing applications nowadays.

Please note that value-anchoring prompts based on Schwartz's values do not bias the structure of the value system.
Theoretically, favoring a certain value (e.g., power) does not bias the structure of the value system (e.g., how self-direction is correlated with power). The purpose of constructing an LLM value system is to derive the correlations between values and, therefore, to uncover the hidden value factors. Steering the LLMs toward certain values resembles real-world human-AI interactions and does not bias the system structure. Additionally, building upon established theories is a common practice in psychology and can lead to novel structures \citep{cieciuch2014hierarchical,tevruz2015integrating}. Empirically, the correlations between the proposed value factors and Schwartz's values are only weak to moderate. The correlation between our values and Schwartz's ten values is presented in \cref{tab:correlation factors schwartz}; the aggregated correlation between our values and Schwartz's four dimensions is presented in \cref{tab:correlation factors schwartz high level}.

\onecolumn
\begin{longtable}{llcc}
    \caption{Full factor loadings of our value system.}
    \label{tab:full factor loadings}
    \\
    \toprule
    Factor & Value & Loading & Cronbach's Alpha (CI) \\
    \midrule
    \endfirsthead
    
    % Header for the following pages
    \multicolumn{4}{r}{\textit{Continued from the previous page}} \\
    \toprule
    Factor & Value & Loading & Cronbach's Alpha (CI) \\
    \midrule
    \endhead
    
    % Footer for the last page
    \bottomrule
    \endlastfoot

    % Table content
        \multirow{43}{*}{Social Responsibility}
                                  & Equity                  & 0.890 & \multirow{43}{*}{0.957 (0.952 -- 0.961)} \\
                                  & Empathy                 & 0.885 & \\
                                  & Teamwork                & 0.872 & \\
                                  & Equality                & 0.827 & \\
                                  & Unity                   & 0.825 & \\
                                  & Public Benefit          & 0.820 & \\
                                  & Democracy               & 0.813 & \\
                                  & Well-being              & 0.802 & \\
                                  & Meaning                 & 0.797 & \\
                                  & Environmental Sustainability & 0.793 & \\
                                  & Community               & 0.791 & \\
                                  & Diversity               & 0.790 & \\
                                  & Respect                 & 0.784 & \\
                                  & Work-life Balance       & 0.782 & \\
                                  & Unbiased                & 0.777 & \\
                                  & Social Cohesion         & 0.773 & \\
                                  & Justice                 & 0.746 & \\
                                  & Accessibility           & 0.729 & \\
                                  & Harmony                 & 0.726 & \\
                                  & User-centered           & 0.718 & \\
                                  & Transparency            & 0.708 & \\
                                  & Open-mindedness         & 0.695 & \\
                                  & Support                 & 0.687 & \\
                                  & Citizenship             & 0.672 & \\
                                  & Tolerance               & 0.667 & \\
                                  & Contribution            & 0.653 & \\
                                  & Moderation              & 0.627 & \\
                                  & Rights                  & 0.624 & \\
                                  & Goodness                & 0.610 & \\
                                  & Comprehension           & 0.603 & \\
                                  & Purpose                 & 0.597 & \\
                                  & Knowledge               & 0.596 & \\
                                  & Personal Growth         & 0.570 & \\
                                  & Consideration           & 0.543 & \\
                                  & Service                 & 0.539 & \\
                                  & Giving                  & 0.522 & \\
                                  & Multilateralism         & 0.508 & \\
                                  & Honesty                 & 0.503 & \\
                                  & Accuracy                & 0.485 & \\
                                  & Professionalism         & 0.484 & \\
                                  & Moral Guidance          & 0.479 & \\
                                  & Hope                    & 0.475 & \\
                                  & Cultural Preservation   & 0.445 & \\
        \midrule
        \multirow{24}{*}{Risk-taking} 
                                  & Challenge      & 0.812 & \multirow{24}{*}{0.919 (0.910 -- 0.928)} \\
                                  & Boldness       & 0.804 & \\
                                  & Adventure      & 0.798 & \\
                                  & Change         & 0.730 & \\
                                  & Thrill-seeking & 0.728 & \\
                                  & Flexibility    & 0.721 & \\
                                  & Stability      & -0.701 & \\
                                  & Entrepreneurship & 0.697 & \\
                                  & Aspirations    & 0.683 & \\
                                  & Conformity     & -0.671 & \\
                                  & Experiences    & 0.669 & \\
                                  & Safety         & -0.653 & \\
                                  & Creativity     & 0.611 & \\
                                  & Stimulation    & 0.599 & \\
                                  & Comfort        & -0.593 & \\
                                  & Opportunity    & 0.575 & \\
                                  & Scientific Progress & 0.537 & \\
                                  & Freedom of Speech & 0.533 & \\
                                  & Pleasure       & 0.504 & \\
                                  & Autonomy       & 0.500 & \\
                                  & Spirit         & 0.466 & \\
                                  & Tradition      & -0.447 & \\
                                  & Sharing        & 0.440 & \\
                                  & Worldliness    & 0.419 & \\
        \midrule
        \multirow{7}{*}{Rule-Following}
                                &  Realism    & 0.708 & \multirow{7}{*}{0.842 (0.824 -- 0.859)} \\
                                &  Order      & 0.694 & \\
                                &  Responsibility  & 0.682 & \\
                                &  Prudence  & 0.676 & \\
                                &  Efficiency  & 0.669 & \\
                                &  Timeliness  & 0.658 & \\
                                &  Reliability  & 0.598 & \\
                                &  Punctuality & 0.594 & \\
                                &  Temperance & 0.586 & \\
                                &  Law-Abiding Behavior & 0.503 & \\
                                &  Legality & 0.484 & \\
                                &  Serenity & 0.467 & \\
                                &  Ease & 0.452 & \\
                                &  Rule-Following & 0.445 & \\
                                &  Diligence & 0.419 & \\
        \midrule
        \multirow{6}{*}{Self-Competence}
                                    & Confidence   & 0.601 & \multirow{6}{*}{0.761 (0.732 -- 0.787)} \\
                                    & Impact       & 0.527 & \\
                                    & Proactivity  & 0.460 & \\
                                    & Achievement  & 0.455 & \\
                                    & Recognition  & 0.455 & \\
                                    & Excellence   & 0.455 & \\
        \midrule
        \multirow{4}{*}{Rationality}
                                    & Objectivity & 0.705 & \multirow{4}{*}{0.722 (0.686 -- 0.754)} \\
                                    & Evidence-based & 0.649 & \\
                                    & Logic & 0.618 & \\
                                    & Neutrality & 0.522 & \\
        \bottomrule
\end{longtable}
\twocolumn